\documentclass[runningheads]{ola}

\usepackage{amsmath}
\usepackage{epsfig}

\begin{document}

\pagestyle{headings}

\mainmatter

\title{Fractal Decomposition based Algorithm for Dynamic camera alignment optimization problem}

% Title
\titlerunning{FDA applied to camera alignment}

% Title for odd pages
\author{Arcadi Llanza\inst{1,2}, Nadiya Shvai\inst{1} \and Amir Nakib\inst{2}}

% Authors for the top of the even pages
\authorrunning{Arcadi Llanza}

\institute{
Vinci Autoroutes, Cyclope.ai, Paris, France \\
\and
Université Paris-Est Créteil, Laboratoire LISSI, Vitry sur Seine, France \\
}

\maketitle

\section{Introduction}

Image Alignment (IA) has become an important task to be taken into account in real-world applications. IA is the process of overlaying images to be able to further analyze them. It is a crucial step because many systems rely on obtaining the correct formatted data to take a posteriory decision. This concept comes from a more generic one called image registration. IA was initially introduced by \emph{Lukas, Kanade et al.} in \cite{lucas1981iterative}. 
% Afterwards, in \cite{baker2004lucas} was presented an overview of IA, describing most of the algorithms and their extensions in a consistent framework.

In this work, we tackle the Dynamic Optimization Problem (DOP) of IA in a real-world application using a Dynamic Optimization Algorithm (DOA) called  Fractal Decomposition Algorithm (FDA), introduced in \cite{nakib2017deterministic} by \emph{Nakib et al.}. We used FDA to perform IA on CCTV camera feed from a tunnel. As the camera viewpoint can change by multiple reasons such as wind, maintenance, etc. the alignment is required to guarantee the correct functioning of video-based traffic security system.

The rest of this paper is organized as follows. First, Section \ref{section:Problem_definition} introduces the problem formulation. Then, Section \ref{section:Methodology} recalls the FDA foundations. Afterwards, \ref{section:Results} introduces the conducted experiments and explains the obtained results. Finally, in Section \ref{section:Conclusions} the conclusions are presented.
\section{Problem Definition}
\label{section:Problem_definition}

\begin{figure}[h]
    \centering
    \includegraphics[width=0.225\textwidth]{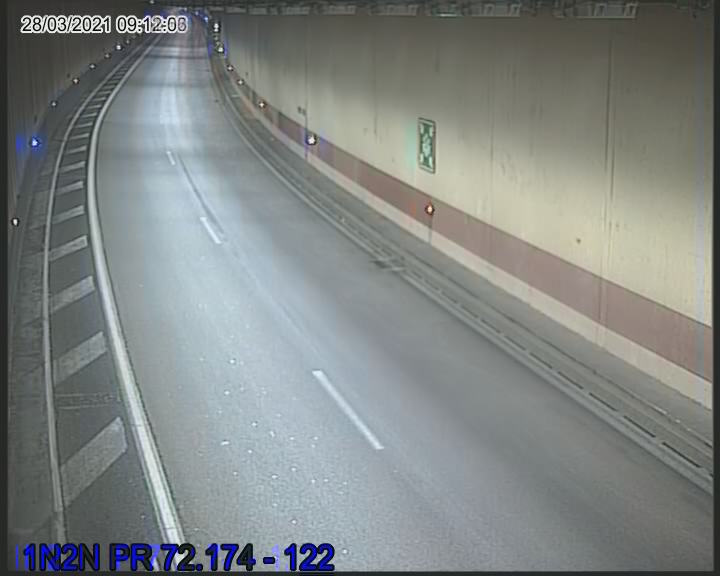}
    \includegraphics[width=0.225\textwidth]{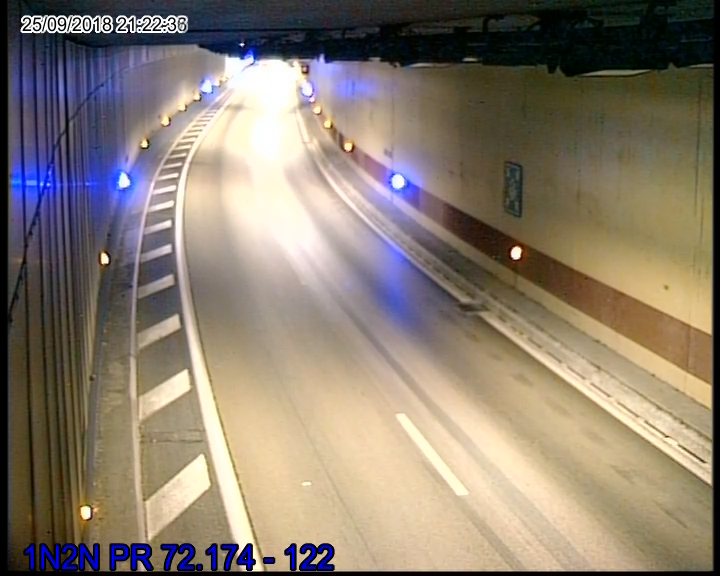}
    \hspace{1cm}
    \includegraphics[width=0.225\textwidth]{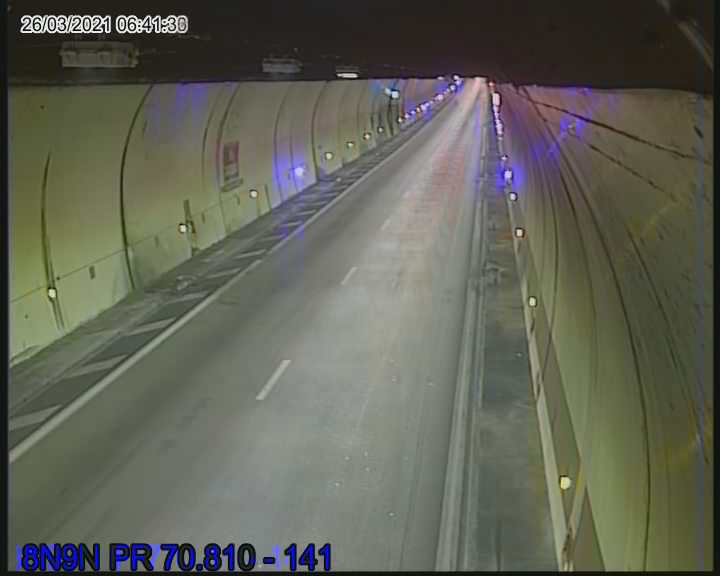}
    \includegraphics[width=0.225\textwidth]{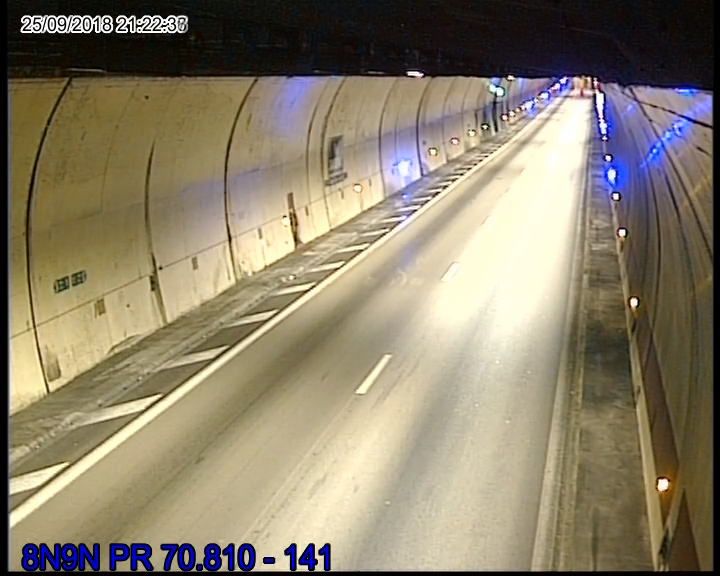}
    \caption{\textbf{Left:} Images obtained from the camera at time $t-1$. \textbf{Right:} Images obtained from the camera at time $t$. A shift in the camera can be observed including translation and rotation distortions.}
    \label{fig:Image_sequence_1}
\end{figure}

In this real-world application, we aim to solve the dynamic optimization problem of aligning a camera image that has been shifting over time (see Figure \ref{fig:Image_sequence_1}).

To do so, we optimize the matrix $H$ (see Equation (\ref{eq:homography})) containing 8 Degrees of Freedom (DoF) that correspond to image geometrical transformation (such as translation, rotation, skewing). 
\begin{equation}
    H=
    \begin{pmatrix}
       DoF_{11} & DoF_{12} & DoF_{13} \\
       DoF_{21} & DoF_{22} & DoF_{23} \\
       DoF_{31} & DoF_{32} & 1 \\
    \end{pmatrix}
    \label{eq:homography}
\end{equation}
The direct and inverse transformation are given in Equation (\ref{eq:transformation}).%allows us to transform an image that is in certain coordinate space into another space coordinate 
\begin{equation}
    \begin{split}
        & I_{1} \cdot H = I_{2} \\
        & I_{1} = I_{2} \cdot H^{-1}
    \end{split}
    \label{eq:transformation}
\end{equation}
Therefore, given two images $I_{1}$ and $I_{2}$ that have been captured using the same camera (from the same position, just changing its viewpoint), and that have a view overlap, it should exist a matrix $H$ that let us transform the content of one image into the other one. Specifically, the values in $H$: $DoF_{11}$, $DoF_{12}$, $DoF_{21}$ and $DoF_{22}$ estimate the rotation. Parameters $DoF_{13}$, $DoF_{23}$ estimate the translation. Finally, $DoF_{31}$, $DoF_{32}$ estimate the skewing effect. Parameter $H_{33}$ is used for the scale magnitude. Given the problem setup it can be fixed to a constant value of 1. 

\begin{figure}[h]
    \centering
    \includegraphics[width=0.45\textwidth]{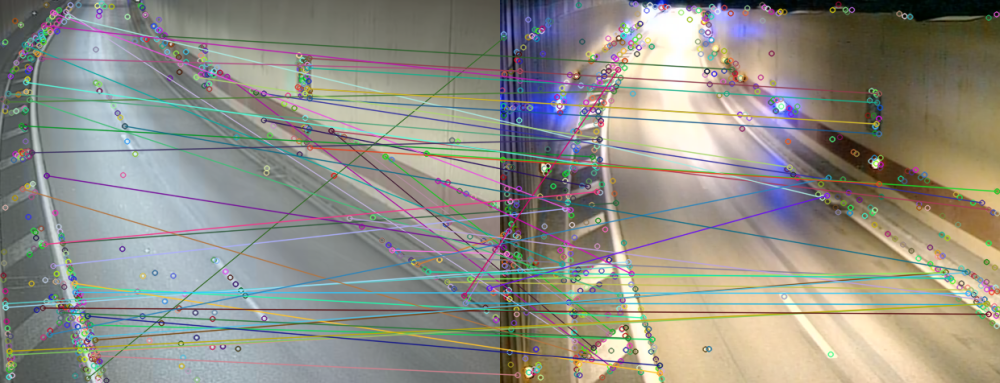}
    \hspace{1cm}
    \includegraphics[width=0.45\textwidth]{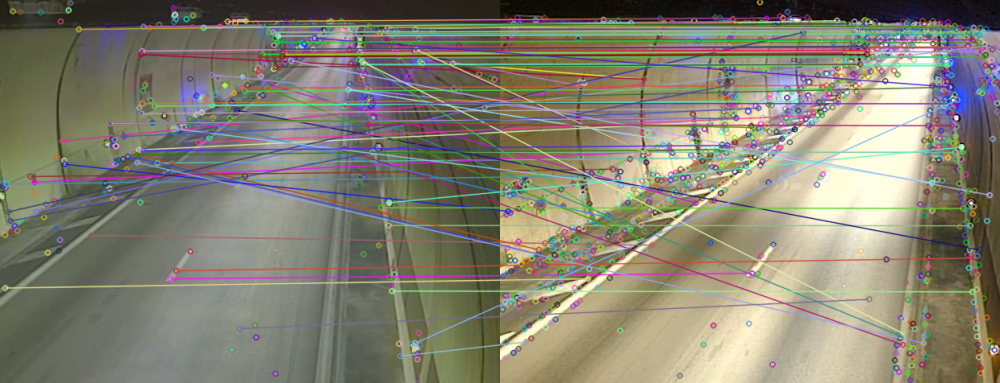}
    \caption{Sets of images containing the detected and matched keypoints.}
    \vspace{-1.5em}
    \label{fig:matched_keypoints}
\end{figure}

To perform estimation % and optimization 
of the $H$ matrix parameters, a set of point have been initially detected and described in both images. These points were then pairwise matched based on the descriptors similarity (see Figure \ref{fig:matched_keypoints} for an illustration).% among them keeping the ones that are close in the features space. 
We formulate the IA as a dynamic optimization problem by considering the loss function $\mathcal{L}_{loss}$ that provides the (partial) sum of distances between pairs of matched keypoints after apply the transformation $H$ on one point from the pair (see Equation (\ref{eq:loss_function})). 
%
%\begin{equation}
%    \begin{split}
%        & \vec{error} = \sum %|(\vec{keypoints_{I_2}} - %\vec{(keypoints_{I_1}} \cdot %H))| \\
%        & loss\_function = \sum %P_{i}(\vec{error})
%    \end{split}
%    \label{eq:loss_function}
%\end{equation}
%
\begin{equation}
    \begin{split}
        & error_k = d_{L_1} \left ( keypoint_{I_2}^k, keypoint_{I_1}^k \cdot H\right ) \\
        &  \mathcal{L}_{loss} = \sum_{error_k < P_{i}({errors})} error_k
    \end{split}
    \label{eq:loss_function}
\end{equation}
More specifically, we use $L_1$ distance between pairs of points. When calculating the loss function we take into account lower the distances up to a certain $i^{th}$ percentile $P_{i}$. For the experiments we found empirically the value of $i=80$ to perform well. We use the percentile to smooth the optimization function (\emph{e.g.} filter non-properly matched keypoints).
\section{Methodology}
\label{section:Methodology}
\begin{figure}[h]
    \centering
    \includegraphics[width=0.3\textwidth]{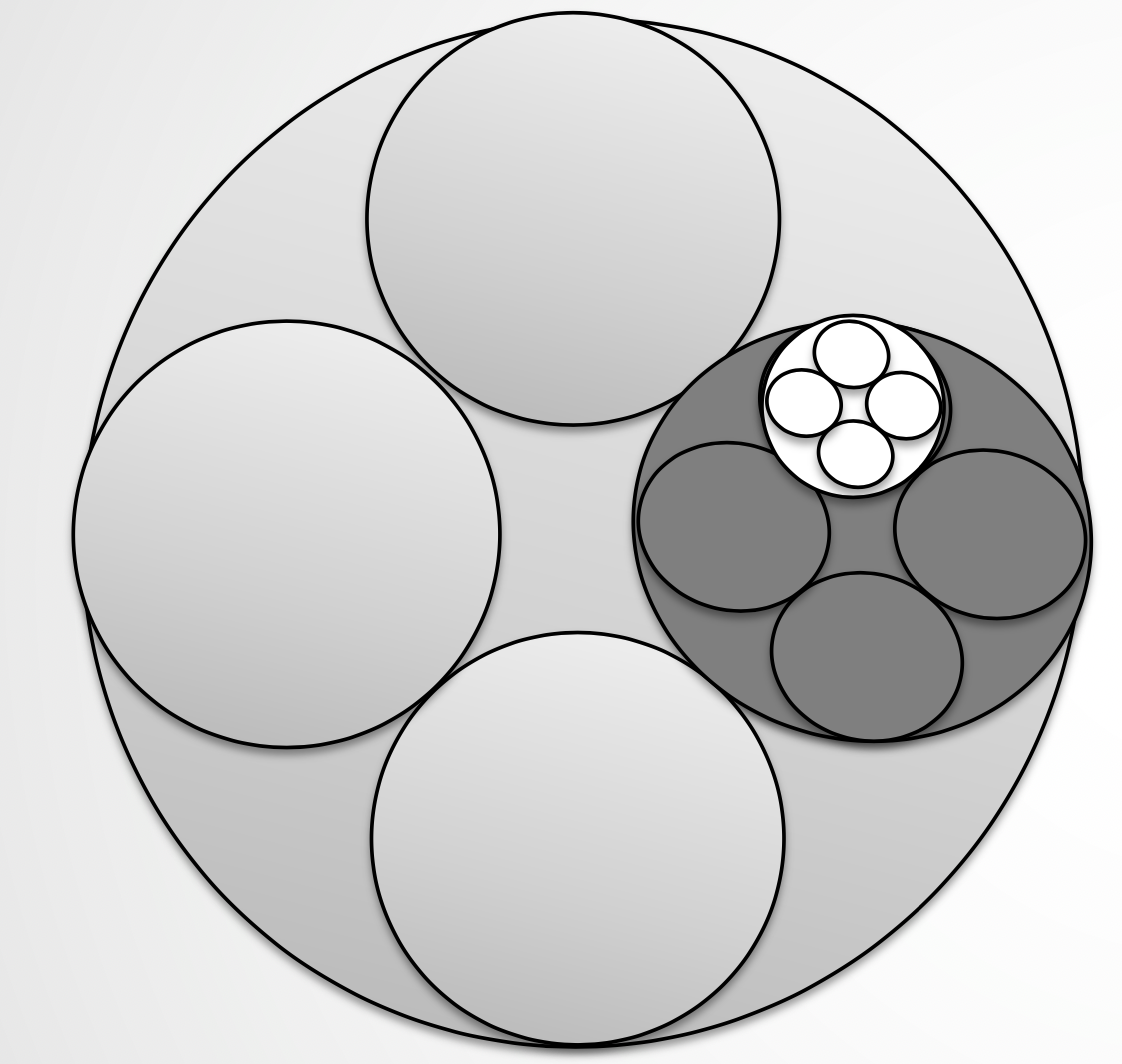}
    \hspace{1cm}
    \includegraphics[width=0.3\textwidth]{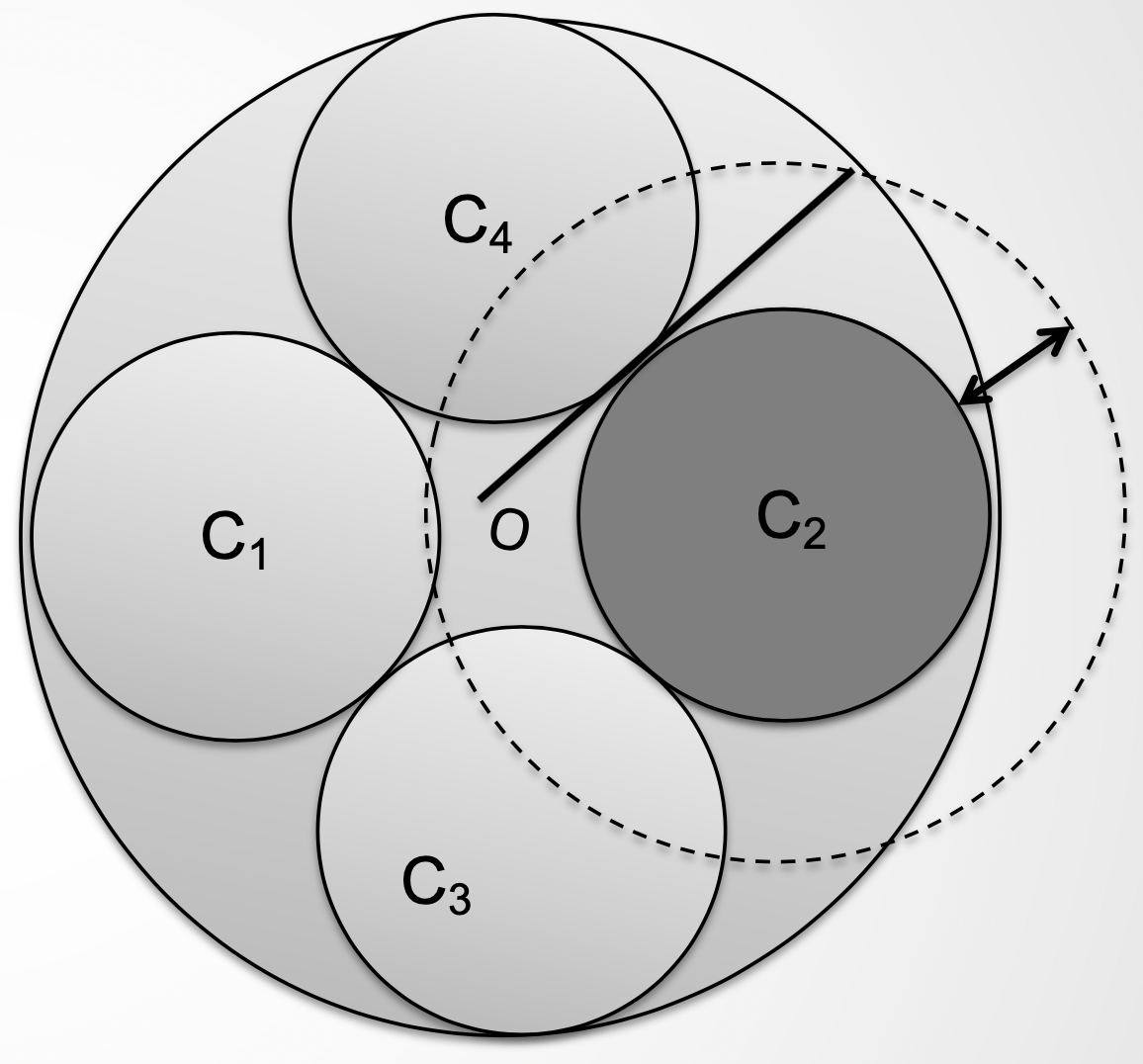}
    \caption{\textbf{Left:} 4-level decomposition using the hypersphere fractal structure. \textbf{Right:} Representation of the hypersphere fractal decomposition with inflation coefficient to ensure that the search space is fully covered given a particular dimension.}
    \vspace{-1.5em}
    \label{fig:FDA_decomposition}
\end{figure}

To solve the stated problem, we propose to use FDA to deal with this dynamic environment.
%DFDA is used to optimize the earlier mentioned case scenario to showcase that it can provide accurate results in a low dimension dynamic optimization problem. 
More specifically, FDA uses a fractal decomposition structure based on hyperspheres (see Figure \ref{fig:FDA_decomposition}) to explore the search space, and an Intensive Local Search (ILS) method to exploit the promising regions. 
It was shown in \cite{souquet2020hyperparameters} that this approach is exceptionally beneficial in high dimensional space problems. Herein, our goal is to demonstrate that FDA can also provide accurate results in a low dimension dynamic optimization problems and can be successfully applied to solve a real-world task.  %optimize successfully in a real-world application that is based on a low dimensional space.
\section{Results}
\label{section:Results}
\begin{figure}[h]
    \centering
    \includegraphics[width=0.5\textwidth]{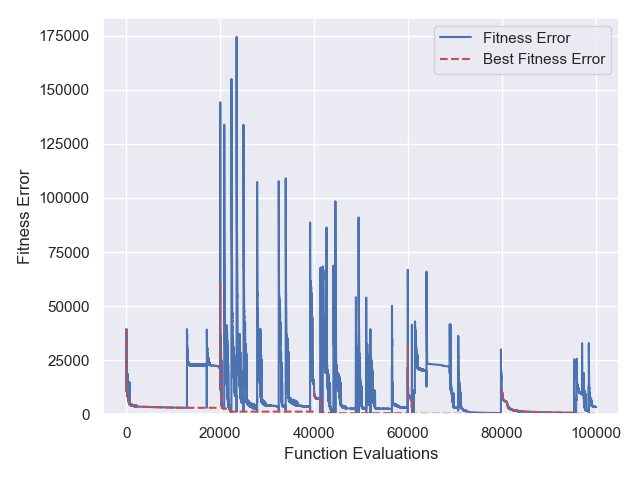}
    \caption{Graph providing the dynamic optimization experiments. In blue the fitness error (obtained from the loss function) can be seen. In red the best fitness error obtained so far during this period of time is shown. Particularly, it can be observed that a camera has moved over time 5 times (red peaks).}
    \vspace{-1.5em}
    \label{fig:loss_function_optimization}
\end{figure}

The Figure \ref{fig:loss_function_optimization} illustrates the optimization process. The red dashed line shows the best fitness error in the current period, and blue line gives the current fitness error obtained at a given evaluation. 

\begin{figure}[h]
    \centering
    \includegraphics[width=0.4\textwidth]{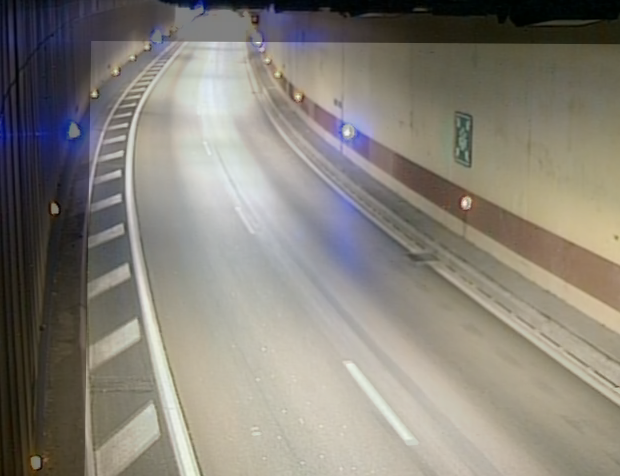}
    \hspace{2cm}
    \includegraphics[width=0.4\textwidth]{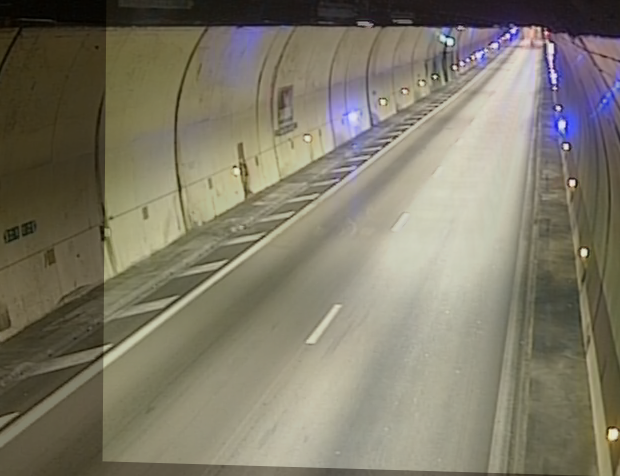}
    \caption{Sets of blended images represented in the same space showing an accurate match.}
    \vspace{-1.5em}
    \label{fig:blended_output}
\end{figure}

In Figure \ref{fig:blended_output} a pair of blended images is given in the same coordinate space showing that FDA has provided accurate results.
\section{Conclusions}
\label{section:Conclusions}

In this paper, we presented a real-life use case of applying FDA. The obtained results demonstrate the efficiency of the method for solving the problem of image alignment. For the future work, we plan to extend FDA  to deal with  higher dimensional real-world problems (\emph{e.g.} computer vision) to further validate the FDA's capabilities. %robustness for solving real-life low dimensional problems such as this, image alignment.

%As an open lines, further tests using higher dimensional real-world problems (\emph{e.g.} computer vision) should be carried on to validate the DFDA's capabilities.

% \begin{thebibliography}{1}
%
% \bibitem{mich}
% Michalewicz, Z.: Genetic Algorithms + Data Structures = Evolution Programs.
% 3rd edn. Springer-Verlag, Berlin Heidelberg New York (1996)
% \end{thebibliography}

\bibliographystyle{IEEEtran}
\bibliography{bibtex/bib/ref}

\end{document}